\documentclass[10pt,twocolumn,letterpaper]{article}

\usepackage{wacv}              % To produce the CAMERA-READY version
\definecolor{Gray}{gray}{0.9}
\definecolor{Red}{RGB}{255, 184, 179}
\definecolor{Green}{RGB}{202, 235, 213}

\newcommand{\cpar}[1]{%
  \begingroup
    \ifthenelse{\lengthtest{#1 pt > 0pt}}{%
      \textcolor{black}{(}\textcolor{OliveGreen}{+#1}\textcolor{black}{)}%
    }{%
      \ifthenelse{\lengthtest{#1 pt = 0pt}}{%
        \textcolor{black}{(}\textcolor{gray}{+0.00}\textcolor{black}{)}%
      }{%
        \textcolor{black}{(}\textcolor{BrickRed}{#1}\textcolor{black}{)}%
      }%
    }%
  \endgroup
}

\newcommand{\cparper}[1]{%
  \begingroup
    \ifthenelse{\lengthtest{#1 pt > 0pt}}{%
      \textcolor{black}{(}\textcolor{OliveGreen}{+#1\%}\textcolor{black}{)}%
    }{%
      \ifthenelse{\lengthtest{#1 pt = 0pt}}{%
        \textcolor{black}{(}\textcolor{gray}{+0\%}\textcolor{black}{)}%
      }{%
        \textcolor{black}{(}\textcolor{red}{#1\%}\textcolor{black}{)}%
      }%
    }%
  \endgroup
}

\definecolor{wacvblue}{rgb}{0.21,0.49,0.74}
\usepackage[pagebackref,breaklinks,colorlinks,allcolors=wacvblue]{hyperref}

\usepackage[capitalize]{cleveref}
\crefname{section}{Sec.}{Secs.}
\Crefname{section}{Section}{Sections}
\Crefname{table}{Table}{Tables}
\crefname{table}{Tab.}{Tabs.}

\usepackage{multirow}
\usepackage{listings}
\usepackage{bbm}
\usepackage{amsmath}
\usepackage{amssymb,algorithm}
\usepackage{algpseudocode}
\usepackage{amssymb}
\usepackage{pifont}
\usepackage[dvipsnames]{xcolor}
\usepackage{ifthen}
\usepackage{cancel}
\usepackage[normalem]{ulem}
\usepackage{colortbl}

\definecolor{codegreen}{rgb}{0,0.6,0}
\definecolor{codegray}{rgb}{0.5,0.5,0.5}
\definecolor{codepurple}{rgb}{0.58,0,0.82}
\definecolor{backcolour}{rgb}{0.95,0.95,0.92}

\newlength\savewidth

\definecolor{softblue}{RGB}{217, 220, 250}

\lstdefinestyle{mystyle}{
    commentstyle=\color{codegreen},
    keywordstyle=\color{blue},
    stringstyle=\color{codepurple},
    basicstyle=\ttfamily\footnotesize,
    breakatwhitespace=false,         
    breaklines=true,                 
    captionpos=b,                    
    keepspaces=true,                 
    showspaces=false,                
    showstringspaces=false,
    showtabs=false,                  
    tabsize=2
}
\def\wacvPaperID{1487} % *** Enter the WACV Paper ID here
\def\confName{WACV}
\def\confYear{2026}

\title{You May Speak Freely: Improving the Fine-Grained Visual Recognition Capabilities of Multimodal Large Language Models with Answer Extraction}

\author{Logan Lawrence$^{1}$ \hspace{.1cm} Oindrila Saha$^{1}$ \hspace{.1cm} Megan Wei$^{2}$ \hspace{.1cm} Chen Sun$^{2}$ \hspace{.1cm} Subhransu Maji$^{1}$ \hspace{.1cm} Grant Van Horn$^{1}$ \\[1ex]
University of Massachusetts, Amherst$^{1}$ \quad Brown University$^{2}$\\[.7ex]
{\tt\small \{lclawrence, osaha, smaji, gvanhorn\}@umass.edu$^{1}$ \quad \{meganwei, chen\_sun4\}@brown.edu$^{2}$}}

\begin{document}

\maketitle

\begin{abstract}

Despite the renewed interest in zero-shot visual classification due to the rise of Multimodal Large Language Models (MLLMs), the problem of evaluating free-form responses of auto-regressive models remains a persistent challenge. Most existing works focus on language-only tasks or don't consider Multiple Choice Questions (MCQs) beyond 5-way options, both of which are critical capabilities to solve tasks in Fine-Grained Visual Classification (FGVC) where choice counts are in the hundreds to thousands and the choices are highly related. Furthermore, in this highly multi-way MCQ setting it is not clear how to extend LLM choice extraction to retrieval-based problems, where computing probabilities over the choice set is computationally costly. In this work we investigate \textit{nlg2choice}, a simple two-stage method which first asks the MLLM an open-ended question for the task with minimal constraints, then uses text-only constrained decoding to predict the most likely choice. In retrieval settings, we compute the probability of the constrained response taking that choice with an early stopping method to significantly improve throughput. Our results show improvement over a suite of seven fine-grained visual datasets when evaluating in terms of classification and retrieval, and show that this performance holds over the various ways that users of LLMs can implement tasks in natural language. 

\end{abstract}

\vspace{-15pt}

\section{Introduction}

% TODOs
% visualization of first token probs
% more templates (dan sheldon idea of "what are all the common ways to respond")
% gull example of decreasing performance
% dump all prompts in appendix
% closed source models

% TODO where it has become common practice to formulate classification datasets as Visual Question Answering (VQA) tasks with Multiple Choice Questions (MCQs) and a common prompt \cite{?}

% Zero-shot visual classification has been revitalized by the introduction of Multimodal Large Language Models (MLLMs) \cite{llava, instructblip, gpt4v, llava_next}, where one can formulate classification datasets as Visual Question Answering (VQA) tasks with Multiple Choice Questions (MCQs) and a common prompt. To deal with the free-form natural language outputs of auto-regressive models, visual classification with MLLMs has borrowed techniques from language-only methods which limit the answer space to a defined set of responses, thereby enabling evaluation using ground-truth labels typically through alphabetical lettering of the possible choices \cite{mmbench}.

\begin{figure}[t!]
    \centering
    \captionsetup{type=figure}
    \setlength{\belowcaptionskip}{-10pt}
    \includegraphics[width=.9\linewidth]{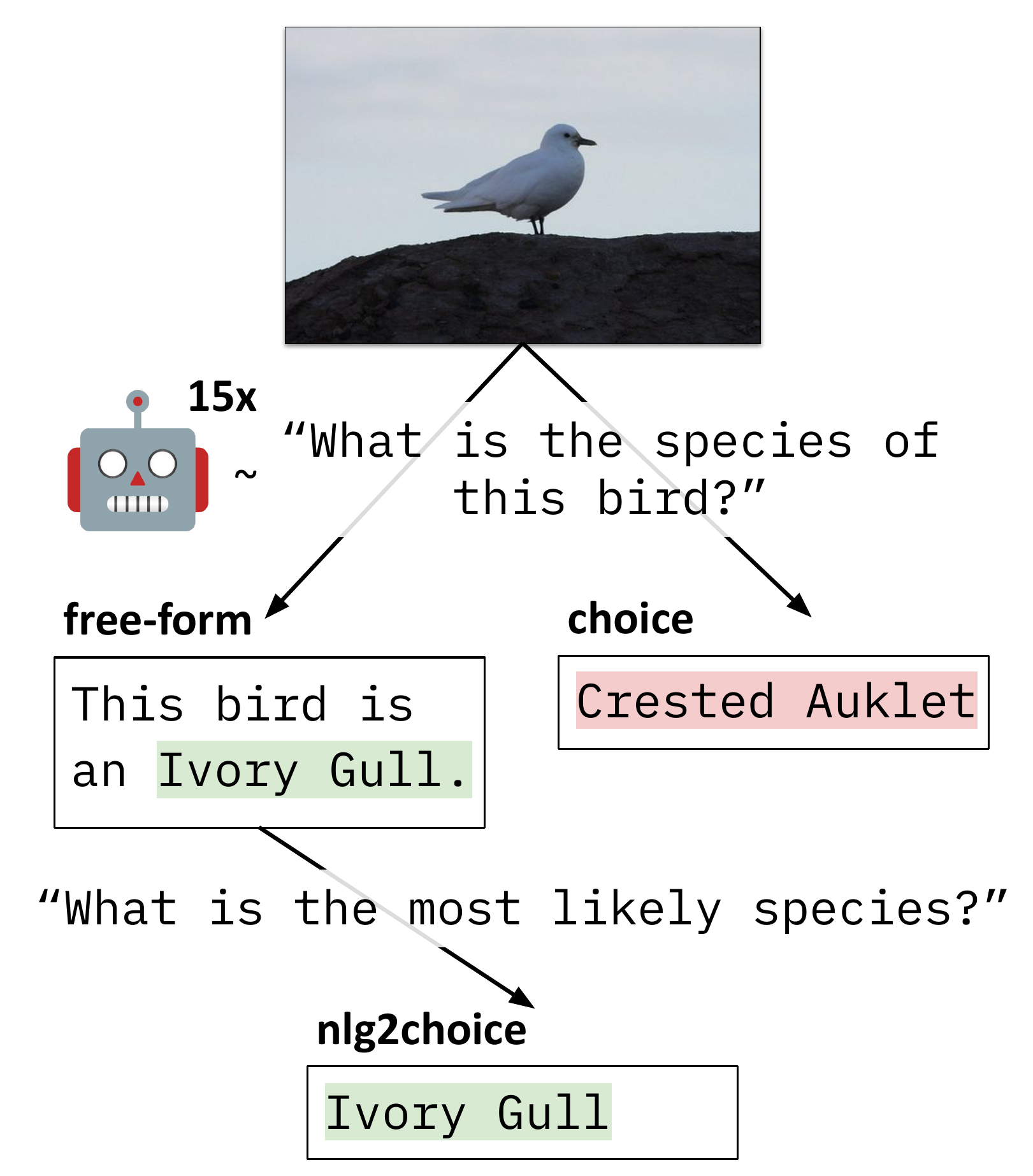}
    \vspace{0pt}
    \captionof{figure}{\textbf{Proposed evaluation setup of \textit{nlg2choice}}. Whereas previous approaches for MLLM classification use methods like probability sequences or constrained decoding, \textit{nlg2choice} first prompts the MLLM for a natural language response, then performs \textit{text-only} constrained decoding on the text to extract classes. Accuracy is computed as an average over multiple rewritings of prompts as a robustness measure. Correct MLLM responses are shown in \colorbox{Green}{green} whereas incorrect responses are \colorbox{Red}{red}.}
    \label{fig:preview_combined}
\end{figure}

% Zero-shot visual classification has been revitalized by the introduction of Multimodal Large Language Models (MLLMs) \cite{llava, instructblip, gpt4v, llava_next}, where one can formulate classification datasets as Visual Question Answering (VQA) tasks with Multiple Choice Questions (MCQs) and a common prompt. However, \textit{fine-grained} visual classification (FGVC) poses a barrier to traditional MCQ problems due to two issues: (1) a \textbf{vastly higher number of choices} (often hundreds to thousands) and (2) the \textbf{niche concepts} that each choice represents.

Zero-shot visual classification has been revitalized by the introduction of Multimodal Large Language Models (MLLMs) \cite{llava, instructblip, gpt4v, llava_next}, where one can formulate classification datasets as Visual Question Answering (VQA) tasks with Multiple Choice Questions (MCQs) and a common prompt. However, \textit{fine-grained} visual classification (FGVC) poses a barrier to traditional MCQ problems due to two issues: (1) a \textbf{vastly higher number of choices} (often hundreds to thousands) and (2) the \textbf{niche concepts} that each choice represents. Recent evaluations have found that state-of-the-art instruction-tuned MLLMs (e.g. LLaVA 1.5 \cite{llava, llava_next}, InstructBLIP \cite{instructblip}, GPT-4V \cite{gpt4v}) \textbf{suffer dramatic drops in accuracy} when asked to name fine-grained categories. For instance, LLaVA-1.5 (13B) achieves near-perfect accuracy on coarse categories (like “bird” vs “not bird”), but only ~1–2\% accuracy on fine-grained species labels in iNaturalist \cite{finer}. 

% In short, MLLMs often fail to distinguish very similar sub-classes or assign choices reliably in many-way scenarios \cite{?}, highlighting the need for more sophisticated guided strategies.

% These problems are fundamental challenges to the most popular MCQ approaches, exacerbating existing issues.

% let me speak freely
% https://arxiv.org/pdf/2505.04016, also has a section on constrained decoding
% outlines teams response to lmsf
% https://www.reddit.com/r/MachineLearning/comments/1gwswn7/r_say_what_you_mean_a_response_to_let_me_speak/

% self consistency
% https://arxiv.org/pdf/2203.11171

\begin{figure*}[ht!]
    \centering
    \setlength{\belowcaptionskip}{-10pt}
    \captionsetup{type=figure}
    \includegraphics[width=\linewidth]{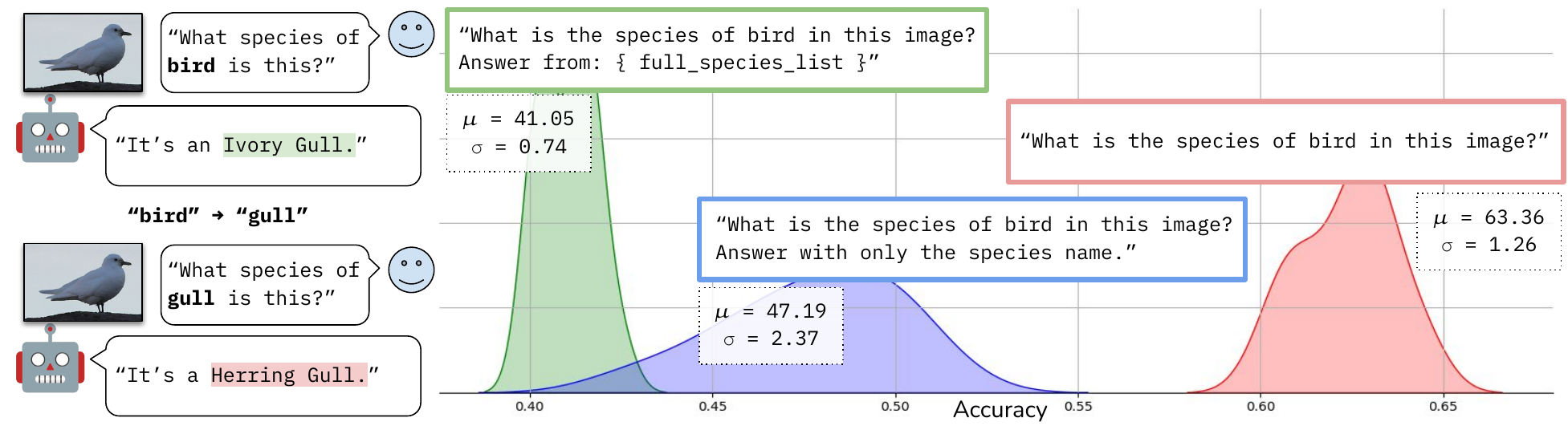}
    \vspace{-20pt}
    \captionof{figure}{\textbf{Seemingly innocuous instruction changes greatly affect MLLM performance.} When testing on the gull subset of CUB200\cite{cub_dataset}, replacing one word can change the predictions of a model \textbf{(left)}. When rewriting an instruction in a semantically equivalent way, performance varies substantially for Qwen-2.5VL \textbf{(right)}. Each distribution comprises 15 semantically equivalent ways of writing the same instruction. These cases highlight the need to test LLMs over various equivalent inputs to ensure robustness.}
    \label{fig:brittle}
\end{figure*}

\textit{Using LLMs to extract choices from pre-existing text} appears to be a potential solution to this problem. Constraining LLM's output format to ease extraction can simplify scoring but might hinder the model’s reasoning \cite{wang2025slot, tam2024let, wang2024look}, whereas allowing free-form explanations improves reasoning at the cost of harder extraction \cite{molfese2025right, tam2024let}. Many approaches prompt a strong LLM \cite{chiang2023vicuna, li2023alpacaeval} such as GPT4 \cite{achiam2023gpt} to perform second-stage answer extraction. However, the works that propose LLMs as choice extractor are either focused only on language-only tasks \cite{lyu2024beyond, wang2024my} or do not extend beyond simple lettering or numbering schemes \cite{mmbench}. Furthermore, no significant treatment of retrieval nor the robustness of these results under input prompt variation has been considered. 

% In this work, we investigate whether evaluating on the NLG directly yields improvements in the visual fine-grained setting. Our contributions are as follows: (1) \textit{nlg2choice}, a simple method to increase FGVC performance of MLLMs by first asking an open-ended question for the task, then using text-only constrained decoding to predict the most likely choice, (2) performance of \textit{nlg2choice} on a suite of popular fine-grained visual recognition benchmarks in the classification and retrieval setting, (3) a small set of labeled MLLM responses for fine-grained answer extraction, and (4) evaluation of various LLMs as answer extractors.

In this work, we conduct a systematic analysis into constrained decoding with LLMs as answer extractors on free-form responses to fine-grained visual recognition tasks. We come to the following conclusions:
\vspace{.15cm}

\noindent
\textbf{1. Answer extraction on free-form responses improves visual recognition capabilities.} We find that performance improves substantially across 7 FGVC benchmarks when viewed from both classification (\cref{table:fgvc_performance}) and retrieval (\cref{table:retrieval_performance}) perspectives. It additionally improves misclassifications, with most responses being within the same genus (\cref{table:genus_performance}). To overcome the computational costs associated with generating probabilities over hundreds to thousands of choices, we propose an early stopping method (\cref{sec:method:retrieval}) which increases throughput substantially (\cref{table:throughput_increase}). 
\vspace{.15cm}

\noindent
\textbf{2. Answer extraction is robust to user variation.} We test robustness by generating additional semantically equivalent ways of writing an instruction (\cref{sec:method:prompt_variation}) and see that performance improvements are statistically significant when viewing user writing as a source of randomness (\cref{fig:statistical_significance}). We also show that instructions which constrict the response decrease performance (\cref{table:constrictive_performance}) and that methods which rely on forcing the LLM to generate some initial tokens to stimulate Chain-of-Thought (CoT) reasoning do not consistently increase performance (\cref{table:cot_performance}).
\vspace{.15cm}

\noindent
\textbf{3. Constrained decoding is a reliable answer extractor without additional training.} We gauge the answer extraction abilities of LLMs out-of-the-box by labeling spans and assigning answers to free-form NLG responses (\cref{sec:appendix:answer_extraction}). We find that the main bottleneck is not answer extraction (\cref{table:extraction_performance}) as studied in previous works, but a lack of extractable content from free-form responses themselves (\cref{fig:extraction_dataset_breakdown}). We make this data publicly available\footnote{\url{https://github.com/Racerx250/nlg2choice}}.

\vspace{-.1cm}

% https://arxiv.org/pdf/2402.14499
% https://arxiv.org/pdf/2305.13264
% https://arxiv.org/abs/2305.04388

\section{Related Work}

\paragraph{Forcing MLLMs to Generate Valid Choices} The earliest works first prompted an LLM directly for a response then did regex parsing to extract choices \cite{wang2022self}, but it was shown that direct open-ended querying often underperforms on fine-grained tasks \cite{finer}. While generating the full probability the text of each choice sequence is as an option, it remains difficult to implement in practice due to its computational cost \cite{robinson2023leveraginglargelanguagemodels}. The most dominant choice probability generation rank the log probabilities assigned to the option IDs (e.g. "A/B/C/D") from the first token prediction of the model. This approach is widely adopted in many different benchmarks and LLM evaluation studies \cite{hendrycks2020measuring, wang2024mmlu, srivastava2022beyond, liang2022holistic, santurkar2023whose}.

% TODO There are many open-source tools that implement this approach \cite{llama_cpp, guidance_ai, dong2024xgrammar},

% got this from SLOT paper
Parallel to this, constrained decoding techniques seek to guarantee consistent outputs by guiding the
generation process during sampling by constricting the output vocabulary \cite{beurer2024guiding, liu2024we}. There are many open-source tools that implement this approach \cite{dong2024xgrammar, GuidanceAI2023, Gerganov2023, outlines} and while these methods ensure model responses lie within the choice set, recent evaluations have shown that they decrease performance \cite{wang2025slot, tam2024let}. Our work sits in opposition to these recent results, where we posit that constrained decoding is advantageous in the fine-grained setting where previous approaches like first token prediction fail.
\vspace{-.1cm}

\paragraph{MLLMs as Answer Extractors} Recent works have shown a mismatch difference between text-based outputs and token-probability metrics of LLMs on language-only tasks, revealing that a model's first-token likelihood often fails to correspond with its generated answer \cite{lyu2024beyond, wang2024my, wang2024look}. Evaluating a model by reading its full text answer instead of only the first-token or forced letter has been shown to improve robustness, reinforcing the value of answer extraction methods \cite{wang2024look}. Many works use large proprietary models like GPT-4 \cite{achiam2023gpt} to extract answers from free-form text \cite{mmbench, chiang2023vicuna, li2023alpacaeval}.

% For MLLMs, the MMBench benchmark \cite{mmbench} explicitly uses an LLM (GPT-4) as a “helper” to convert open-ended vision-language model outputs into a choice label \cite{mmbench}, showing that ChatGPT-involved extraction highly effective, validating GPT-4’s reliability as a choice extractor. In practice, some open-source evaluation pipelines now include a “judge LLM” module for answer parsing (falling back to exact-match rules otherwise), following a similar idea as MMBench’s approach \cite{chiang2023vicuna, li2023alpacaeval}.

Some recent methods have addressed the problem of answer extraction directly. xFinder \cite{yu2024xfinder} propose a lightweight model fine-tuned specifically to map an LLM’s free-form answer to the correct choice, reporting that their method outperforms a GPT-4 based extractor struggled in terms of extraction accuracy. Similarly, SLOT \cite{wang2025slot} proposes a model-agnostic, lightweight “Structured LLM Output Transformer” that is fine-tuned to post-process any LLM’s free-form text into JSON that exactly matches an arbitrary schema, eliminating the need to modify or constrain the generating model itself. xVerify \cite{xVerify} is an efficient answer verifier for reasoning model evaluations that extracts final answers from complex, multi-step LLM outputs and performs robust equivalence checking—trained on the VAR dataset of multi-model, multi-round–annotated question–answer pairs and achieving over 95\% accuracy and F1, demonstrating strong generalizability. Contrary to these approaches the answer extraction method proposed in this work requires no additional training. We argue that answer extraction is \textit{not} a critical bottleneck when using constrained decoding for fine-grained visual classification and show that LLMs perform decent answer extraction out-of-the-box on popular benchmarks. 

% However, we provide an ablation using these methods as the choice extractor in \cref{sec:results:choice_ablation}.

% The authors introduce a synthetic-data pipeline and dual metrics (schema accuracy and content similarity), showing that a Mistral-7B version of SLOT with constrained decoding achieves 99.5\% schema accuracy and 94\% content fidelity. Even a tiny Llama-3.2-1B paired with SLOT matches or outperforms much larger proprietary systems such as Claude-3.5-Sonnet, demonstrating reliable structured generation in resource-constrained settings. xFinder (2024) developed a dedicated key-answer extraction model, noting that GPT-4 + manual checking is very accurate but impractical for routine use \TODO{cite xFinder}, and demonstrating xFinder’s superiority over regex and GPT-4 on extraction accuracy.

\section{Methodology}

\subsection{Simulating User Variation}\label{sec:method:prompt_variation}
Given the brittleness of LLMs to prompt variations (\cref{fig:brittle}), we wish to test the robustness of LLMs as choice extractors by generating a range of semantically equivalent writings of prompts for a given task. We do this by starting with a small set of hand-written templates for a classification task. Letting \texttt{[choice\_list]} be the list of classes for a given dataset separated by newlines, we define \texttt{[template$_i$]} to be equal to the following:

\definecolor{quotecolor}{rgb}{0.87, 0.95, 0.96}
\begin{quote}
    \makebox[\linewidth]{%
        \colorbox{quotecolor}{%
            \hspace*{0mm} % Adjust left alignment
            \begin{minipage}{\dimexpr\linewidth+10\fboxsep\relax} % Adjust width
                \fontsize{9pt}{10pt}\selectfont % Font settings
                \textbf{(Template 1)} "What is the species of bird in this image?" \\[2ex]
                \textbf{(Template 2)} "What is the species of bird in this image? Answer only with species name and no other text." \\[2ex]
                \textbf{(Template 3)} "What is the species of bird in this image? Answer only from the following list with no other text: \texttt{[choice\_list]}"
            \end{minipage}%
        }%
    }
\end{quote}

Where the words "species" and "bird" are updated to suite the benchmark dataset. Given \texttt{[template$_i$]}, we use o3-high \cite{o3} to generate 14 additional prompt variations:

\definecolor{quotecolor}{rgb}{0.87, 0.95, 0.96}
\begin{quote}
    \makebox[\linewidth]{%
        \colorbox{quotecolor}{%
            \hspace*{0mm} % Adjust left alignment
            \begin{minipage}{\dimexpr\linewidth+10\fboxsep\relax} % Adjust width
                \fontsize{9pt}{10pt}\selectfont % Font settings
                \textbf{(Generate Variations)} What are 14 semantically equivalent ways of writing the following question? \\[2ex]      
                Prompt: \texttt{[template$_i$]}
            \end{minipage}%
        }%
    }
\end{quote}

Next, we perform a manual check on this set of prompts for semantic equivalence to the input prompt, resulting in a total of $45$ variations for each dataset. For the remainder of the paper, accuracies reported are across these variations. For the full list of these prompts, see \cref{sec:appendix:prompt_varations}.

\subsection{\textit{nlg2choice} for Classification} 
% \subsection{Prediction with \textit{nlg2choice}} 

% \input{ICCV2025-Author-Kit-Feb/old/method_classification}

Our approach to classification involves two stages: (1) prompting the LLM for a free-form response to a question defined by \cref{sec:method:prompt_variation} and (2) prompting the same model again to select the most likely species given the previous response. On the second step, we enforce a valid output by using \textit{constrained decoding across the choice set}. Additionally, the second stage is text-only, namely it does not see the image which produced the original response. It receives the following prompt:

\definecolor{quotecolor}{rgb}{0.87, 0.95, 0.96}
\begin{quote}
    \makebox[\linewidth]{%
        \colorbox{quotecolor}{%
            \hspace*{0mm} % Adjust left alignment
            \begin{minipage}{\dimexpr\linewidth+10\fboxsep\relax} % Adjust width
                \fontsize{9pt}{10pt}\selectfont % Font settings
                \textbf{(Text-Only Choice)} What is the most likely species of bird indicated in this response? \\[2ex]      
                Response: \texttt{[nlg]} \\[2ex]
                Answer from the following: \texttt{[choice\_list]}
            \end{minipage}%
        }%
    }
\end{quote}
Where \texttt{[nlg]} is the free-form response to the dataset instruction and \texttt{[choice\_list]} is newline-separated list of class names. Similarly, "species" and "bird" are updated accordingly for the given dataset.

\subsection{\textit{nlg2choice} for Retrieval} \label{sec:method:retrieval}

In thresholding-based scenarios one desires a continuous score for every query. Most commonly this is done by taking the probability of \texttt{"Yes"} and \texttt{"No"} for each image, class pair $\mathbf{x}_i \in \mathcal{X}, y_i \in \mathcal{Y}$, ie.:
\begin{align*}
    \hat{y}_i = \text{softmax}(
        \begin{bmatrix}
        p(\texttt{"Yes"}|\mathbf{x}_i, \texttt{[prompt]}) \\
        p(\texttt{"No"}|\mathbf{x}_i, \texttt{[prompt]})
    \end{bmatrix}
\end{align*}
Where $p(\cdot)$ is the next-token probability distribution of a MLLM and \texttt{[prompt]} describes a two-way prediction task on one class of the dataset, eg. \texttt{"Is this a [cname]?"} and \texttt{[cname]} is the name of some class. However, this requires $|\mathcal{X}|\cdot|\mathcal{Y|}$ forward passes of the model. 

% \begin{figure}[t]
%     \centering
%     \captionsetup{type=figure}
%     \includegraphics[width=\linewidth]{ICCV2025-Author-Kit-Feb/placeholders/truncated_prob_1.png}
%     \vspace{0pt}
%     \captionof{table}{\textbf{Number of forward passes needed for different methods in the retrieval setting of \textit{nlg2choice.}} "Full Prob" refers to the probability of the full class name $p(\texttt{[cname]})$ under the LLM, where \texttt{[cname]} is a common name of a class. "Yes"/"No" refers to the common $|\mathcal{X}|\cdot|\mathcal{Y}|$ method of calculating probability over all the classes $y_i \in \mathcal{Y}$. "Truncated Prob" refers to the process outlined in \cref{sec:method:retrieval}. The percentages in the "Yes"/"No" and "Truncated Prob" columns refer to the throughput increase over caculating the full sequence probabilities "Full Prob." Every setting for truncation is faster than "Yes"/"No" except for one benchmark - NABirds \cite{nabirds_dataset}.}
%     \label{fig:preview_combined}
% \end{figure}%

\begin{table}[t]
\setlength{\belowcaptionskip}{-20pt}
\setlength{\tabcolsep}{9pt}
\begin{center}
{\footnotesize
\begin{tabular}{lccc}
\hline
 \\[-2ex]
\textbf{Method} & \textbf{Full Prob} & \textbf{Yes/No} & \textbf{Truncated Prob} \\
 \\[-2ex]
\hline
 \\[-2ex]
CUB200 & 687 & 200 & \textbf{47} \\ 
~ & \cparper{0} & \cparper{244} & \textbf{\cparper{1362}} \\
Flowers & 257 & 102 & \textbf{12} \\ 
~ & \cparper{0} & \cparper{152} & \textbf{\cparper{2042}} \\
Aircrafts & 479 & 100 & \textbf{79} \\ 
~ & \cparper{0} & \cparper{379} & \textbf{\cparper{506}} \\ 
Cars & 1668 & 196 & \textbf{125} \\ 
~ & \cparper{0} & \cparper{751} & \textbf{\cparper{1234}} \\ 
Food & 269 & 101 & \textbf{24} \\ 
~ & \cparper{0} & \cparper{166} & \textbf{\cparper{1021}} \\ 
NABirds & 3294 & \textbf{555} & 701 \\ 
~ & \cparper{0} & \textbf{\cparper{494}} & \cparper{370} \\ 
iNat-Birds & 5450 & 1486 & \textbf{480} \\ 
~ & \cparper{0} & \cparper{267} & \textbf{\cparper{1035}} \\ 
 \\[-2ex]
\hline
\end{tabular}
\captionof{table}{\textbf{Number of forward passes needed for different methods in the second-stage answer extraction of \textit{nlg2choice.}} "Full Prob" refers to the probability of a full class name $p(\texttt{[cname]})$ under the LLM, where \texttt{[cname]} is a common name of a class. ``Yes/No" refers to the common $|\mathcal{X}|\cdot|\mathcal{Y}|$ method of calculating probability over all the classes $y_i \in \mathcal{Y}$. ``Truncated Prob" refers to the process outlined in \cref{sec:method:retrieval}. The percentages in the ``Yes/No" and "Truncated Prob" columns refer to the throughput increase over caculating the full sequence probabilities "Full Prob." Every setting for truncation is faster than ``Yes/No" except for one benchmark - NABirds \cite{nabirds_dataset}. The fastest method for each benchmark is \textbf{bolded}.}
% \captionof{table}{\textbf{Number of forward passes needed for different methods in the second-stage answer extraction of \textit{nlg2choice.}} "Full Prob" refers to the probability of the full class name $p(\texttt{[cname]})$ under the LLM, where \texttt{[cname]} is a common name of a class. ``Yes/No" refers to the common $|\mathcal{X}|\cdot|\mathcal{Y}|$ method of calculating probability over all the classes $y_i \in \mathcal{Y}$. ``Truncated Prob" refers to the process outlined in \cref{sec:method:retrieval}. The percentages in the ``Yes/No" and "Truncated Prob" columns refer to the throughput increase over caculating the full sequence probabilities "Full Prob." Every setting for truncation is faster than ``Yes/No" except for one benchmark - NABirds \cite{nabirds_dataset}. The fastest method for each benchmark is \textbf{bolded}.}
\label{table:throughput_increase}
}
\end{center}
\end{table}

% \vspace{-.5cm}

To improve throughput, we propose to update the second step of \textit{nlg2choice} to stop generating probabilities when the remaining tokens for a given choice do not occur in any other choices. For example, take the case of \texttt{"Baltimore Oriole"} which can be decomposed into \texttt{["B", "altimore", " Ori", "ole"]}. We can see that in the CUB200 species list there are many names which share the first token \texttt{"B"}, ie. \texttt{"Bewick Wren"} or \texttt{"Belted Kingfisher"}. However, \texttt{"Baltimore Oriole"} is the only choice name which has \texttt{"altimore"} as its second token. Thus, the probabilities for \texttt{[" Ori", "ole"]} are not calculated. In terms of probabilities, this is equivalent to:
\begin{align*}
    &p_{full}(\texttt{"Baltimore Oriole"}) = p(\texttt{"B"}) \label{eq:full} \tag{1}\\ 
    &\hspace{2cm} \times p(\texttt{"altimore"}|\texttt{"B"}) \\
    &\hspace{2cm} \times p(\texttt{" Ori"}|\texttt{"Baltimore"}) \\
    &\hspace{2cm} \times p(\texttt{"ole"}|\texttt{"Baltimore Ori"}) \\[2ex]
    &p_{trunc}(\texttt{"Baltimore Oriole"}) = p(\texttt{"B"}) \label{eq:trunc} \tag{2}\\
    &\hspace{2cm} \times p(\texttt{"altimore"}|\texttt{"B"}) \\
    &\hspace{2cm} \times \textcolor{gray}{\cancel{p(\texttt{" Ori"}|\texttt{"Baltimore"})}} \\
    &\hspace{2cm} \times \textcolor{gray}{\cancel{p(\texttt{"ole"}|\texttt{"Baltimore Ori"})}} 
\end{align*}
On many datasets, this change produces substantial throughput increase as many choice names can easily be determined by their first or second tokens. The special property of truncated probability sequences is that \textit{they have the same probabilities as sampling with constrained generation}. This throughput increase is depicted in \cref{table:throughput_increase}. 

% The special property of truncated probability sequences is that \textit{they have the same probabilities as sampling with constrained generation} in the classification setting. This is due to the fact that once enough of a choice has been generated that doesn't overlap with any other choices, the probability of the remaining tokens in that choice is 1. 

\section{Experiments}

\paragraph{Datasets} We conduct experiments over a benchmark of 7 FGVC datasets: \textbf{CUB} \cite{cub_dataset} (200 classes), \textbf{Flowers
102} \cite{flowers_dataset} (102 classes), \textbf{Stanford Cars} \cite{cars_dataset} (196 classes),
\textbf{FGVC Aircrafts} \cite{aircrafts_dataset} (100 classes) and \textbf{Food101} \cite{food_dataset} (101
classes), \textbf{NABirds} \cite{nabirds_dataset} (555 classes), and \textbf{iNaturalist-Birds} \cite{inat_dataset} (1486 classes). We choose these benchmarks because they are well-known for fine-grained visual classification and are traditionally difficult for zero-shot visual models to solve. iNaturalist-Birds is the validation split of the iNat2021 dataset where the the "class" field of each row is equal to "Aves," resulting in a test size of $14960$. For each dataset we use the full test set and the training split remains unused.  

\vspace{.3cm}
\noindent
\textbf{Evaluation Metrics} \hspace{.1cm} For the classification setting we use accuracy averaged over 15 semantically equivalent prompts, whereas for the retrieval setting we use Mean Average Precision (mAP). Choice lists are the names of classes themselves rather than letters or numbers. As compared to previous works \cite{tan2025vision}, we perform the full-way class prediction rather than subsetting the choices to a 4-way prediction. However, we provide an experiment evaluating this setting in \cref{sec:results:4_way_vqa}.

\begin{table*}[t]
\setlength{\belowcaptionskip}{-20pt}
\setlength{\tabcolsep}{11pt}
\begin{center}
{\footnotesize
\begin{tabular}{lcccccccc}
\hline
 \\[-2ex]
\textbf{Method} & \textbf{Birds} & \textbf{Flowers} & \textbf{Aircrafts} & \textbf{Cars} & \textbf{Foods} & \textbf{NABirds} & \textbf{iNat-Birds} & \textbf{Average}\\
 \\[-2ex]
\hline
\rowcolor{Gray} &&&&&&&& \\[-1.8ex]
\rowcolor{Gray}
\multicolumn{9}{l}{Qwen-2.5VL-7B} \\[.3ex]
 \\[-2ex]
choice & 40.08 & 51.29 & 44.97 & 26.80 & 57.93 & 37.77 & 17.29 & 39.45 \\[.7ex]
nlg2choice & 47.72 & 63.22 & 59.96 & 37.52 & 68.99 & 44.88 & 21.47 & 49.11 \\ 
~ & \cpar{7.64} & \cpar{11.93} & \cpar{14.99} & \cpar{10.72} & \cpar{11.06} & \cpar{7.11} & \cpar{4.18} & \cpar{9.66} \\[.7ex]
nlg2choice$_{open}$ & \textbf{59.88} & \textbf{78.03} & \textbf{61.54} & \textbf{47.13} & \textbf{72.26} & \textbf{50.19} & \textbf{25.86} & \textbf{56.41} \\ 
~ & \textbf{\cpar{19.80}} & \textbf{\cpar{26.74}} & \textbf{\cpar{16.57}} & \textbf{\cpar{20.33}} & \textbf{\cpar{14.33}} & \textbf{\cpar{12.42}} & \textbf{\cpar{8.57}} & \textbf{\cpar{16.97}} \\
 \\[-2ex]
\hline
\rowcolor{Gray} &&&&&&&& \\[-1.8ex]
\rowcolor{Gray}
\multicolumn{9}{l}{Llama-3.2-Vision-11B} \\[.3ex]
 \\[-2ex]
choice & 42.27 & 45.68 & 31.38 & 24.68 & 64.70 & 36.69 & 13.82 & 37.03 \\[.7ex]
nlg2choice & 46.96 & 51.16 & 30.75 & 32.72 & 64.52 & 41.34 & \textbf{15.81} & 40.47 \\ 
~ & \cpar{4.69} & \cpar{5.48} & \cpar{-0.63} & \cpar{8.04} & \cpar{-0.18} & \cpar{4.65} & \textbf{\cpar{1.99}} & \cpar{3.43} \\[.7ex]
nlg2choice$_{open}$ & \textbf{49.59} & \textbf{62.24} & \textbf{37.12} & \textbf{36.78} & \textbf{72.72} & \textbf{46.37} & 13.85 & \textbf{45.52} \\ 
~ & \textbf{\cpar{7.32}} & \textbf{\cpar{16.56}} & \textbf{\cpar{5.74}} & \textbf{\cpar{12.10}} & \textbf{\cpar{8.02}} & \textbf{\cpar{9.68}} & \cpar{0.03} & \textbf{\cpar{8.49}} \\ 
 \\[-2ex]
\hline
\rowcolor{Gray} &&&&&&&& \\[-1.8ex]
\rowcolor{Gray}
\multicolumn{9}{l}{Intern3VL-8B} \\[.3ex]
 \\[-2ex]
choice & 11.27 & 24.68 & 15.84 & 14.78 & 44.12 & 6.90 & 2.28 & 17.12 \\[.7ex]
nlg2choice & 13.81 & 32.52 & 18.63 & 18.34 & 43.36 & 7.06 & 1.95 & 19.38 \\ 
~ & \cpar{2.54} & \cpar{7.84} & \cpar{2.79} & \cpar{3.56} & \cpar{-0.76} & \cpar{0.16} & \cpar{-0.33} & \cpar{2.26} \\[.7ex]
nlg2choice$_{open}$ & \textbf{18.47} & \textbf{37.50} & \textbf{21.69} & \textbf{22.15} & \textbf{51.48} & \textbf{12.39} & \textbf{4.29} & \textbf{24.00} \\ 
~ & \textbf{\cpar{7.20}} & \textbf{\cpar{12.82}} & \textbf{\cpar{5.85}} & \textbf{\cpar{7.37}} & \textbf{\cpar{7.36}} & \textbf{\cpar{5.49}} & \textbf{\cpar{2.01}} & \textbf{\cpar{6.87}} \\ 
 \\[-2ex]
\hline
\end{tabular}
\captionof{table}{\textbf{Performance across fine-grained visual classification (FGVC) benchmarks and architectures.} The ``Birds" dataset refers to CUB200. The first row of each model refers to using constrained decoding directly on the model across the class list of each dataset using the prompt in the header. ``choice" is performing constrained decoding on the prompt \texttt{"What is the \{type\} of \{domain\} in this image? Answer with \{type\} only."} ``nlg2choice" refers to letting the model freely respond to the prompt, then using constrained decoding in a text-only approach. ``nlg2choice$_{open}$" refers to removing the appended response-style instruction, eg. \texttt{"Answer with \texttt{\{type\}} only."} All models are their instructed-tuned version. Best performance is \textbf{bolded} for each model.}
\label{table:fgvc_performance}
}
\end{center}
\end{table*}
% https://tex.stackexchange.com/questions/57418/crop-an-inserted-image
% \begin{figure}[t]
%     \centering
%     \captionsetup{type=figure}
%     \includegraphics[width=\linewidth]{ICCV2025-Author-Kit-Feb/placeholders/genus_matches.pdf}
%     \vspace{0pt}
%     \captionof{table}{\textbf{Accuracy of Genus-Level Predictions on Misclassified Examples.} "Genus Matches" refers to the accuracy of model with respect to the translating each species to its corresponding genus. "\% of Data" refers to the amount of data misclassified by the model and is equivalent to $1 - accuracy$ in \cref{table:fgvc_performance}.}
%     \label{table:genus_performance}
% \end{figure}%

\begin{table}[t]
\setlength{\belowcaptionskip}{-25pt}
\setlength{\tabcolsep}{12pt}
\begin{center}
{\footnotesize
\begin{tabular}{lcc}
\hline
 \\[-2ex]
\textbf{Method} & \textbf{Genus Matches} & \textbf{\% of Data} \\
 \\[-2ex]
\hline
\rowcolor{Gray} && \\[-1.8ex]
\rowcolor{Gray}
\multicolumn{3}{l}{Qwen-2.5VL-7B} \\[.3ex]
 \\[-2ex]
choice & 45.99 & 59.92 \\ 
nlg2choice & 56.57 \cpar{10.56} & 52.28 \\ 
nlg2choice$_{open}$ & \textbf{63.90 \cpar{17.91}} & \textbf{40.12} \\ 
 \\[-2ex]
\hline
\rowcolor{Gray} && \\[-1.8ex]
\rowcolor{Gray}
\multicolumn{3}{l}{Llama-3.2-Vision-11B} \\[.3ex]
 \\[-2ex]
choice & 42.52 & 57.73 \\ 
nlg2choice & 61.71 \cpar{19.19} & 53.04 \\ 
nlg2choice$_{open}$ & \textbf{66.37 \cpar{23.85}} & \textbf{50.94} \\ 
 \\[-2ex]
\hline
\rowcolor{Gray} && \\[-1.8ex]
\rowcolor{Gray}
\multicolumn{3}{l}{Intern3VL-8B} \\[.3ex]
 \\[-2ex]
choice & 36.77 & 88.73 \\ 
nlg2choice & 48.79 \cpar{12.02} & 86.19 \\ 
nlg2choice$_{open}$ & \textbf{60.36 \cpar{23.59}} & \textbf{81.53} \\ 
 \\[-2ex]
\hline
\end{tabular}
\captionof{table}{\textbf{Accuracy of genus-level predictions on misclassified examples.} "Genus Matches" refers to the accuracy of model with respect to the translating each species to its corresponding genus. "\% of Data" refers to the amount of data misclassified by the model and is equivalent to $1 - accuracy$ in \cref{table:fgvc_performance}. The species to genus mapping manually created.}
\label{table:genus_performance}
}
\end{center}
\end{table}

\vspace{.3cm}
\noindent
\textbf{Implementation Details} \hspace{.1cm} Constrained decoding is implemented using the Outlines\footnote{\url{https://github.com/dottxt-ai/outlines}} library \cite{outlines} and classes for each setting are detailed in \cref{sec:appendix:prompt_varations}. We start by generating 15 prompts  described in \cref{sec:method:prompt_variation}, then replacing the key features like domain ("birds," "flowers," etc.) and class types ("species," "variant," etc.) for each dataset. This allows us to compare directly between questions for a given model or dataset, and we also test the performance of appending class lists and common CoT methods. For our experiments, we test only open-source models. Specifically, we test the Qwen-2.5VL \cite{qwen25vl}, Llama-3.2 \cite{llama3}, and InternVL3  \cite{internvl3} architectures. For hyperparameters, we use the recommended generation parameters as defined by the HuggingFace card and take max new tokens to be 512. Experiments were ran in a heterogeneous environment of single-GPU nodes. All nodes required a GPU with at least 23GB VRAM, 16 CPU cores, and 64GB RAM. 

\section{Results}

\subsection{Zero-Shot Classification Performance}\label{sec:results:classification}

\paragraph{Answer extraction is better than directly answering.} In \cref{table:fgvc_performance} we report the performance of \textit{nlg2choice} across various architectures and datasets. All architectures enjoyed a bump in performance across individual datasets and on average (\textbf{+16.97}, \textbf{+8.49}, \textbf{+6.87} across Qwen-2.5VL, Llama-3.2V, and Intern3), but model-specific differences can be clearly seen. Likewise, we find that removing explicit answer formatting from the model (changing from ``nlg2choice" and ``nlg2choice$_{open}$"). We find the Aves subset of iNaturalist \cite{inat_dataset} to be the hardest among all the benchmarks, resulting in the smallest increase in performance across all models.

For model-specific differences, we find that Qwen-2.5VL performs the best on average using \textit{nlg2choice}, having an average accuracy of \textbf{56.91} across the 7 datasets. We also find that Qwen-2.5VL experiences the greatest gain, having an average accuracy difference of \textbf{+16.97} from the base constrained decoding method to nlg2choice$_{open}$. Comparatively, we see that Intern3-VL performs the worst across architectures, having the overall smallest increase when using nlg2choice$_{open}$ (\textbf{+6.87}) and the lowest performance (\textbf{24.00}). However, we do still see improvements over all benchmarks.

\vspace{-.3cm}

\paragraph{Misclassifications are more aligned with ground truth.} A natural question is to ask how the classifications qualitatively change under \textit{nlg2choice}. For an example, take \cref{fig:preview_combined} where the ground truth label is \texttt{"Ivory Gull"}: under constrained decoding the model responds with \texttt{"Crested Auklet"} whereas a higher quality misclassificaiton would be \texttt{"Herring Gull"}, \texttt{"California Gull"}, ..., ie. a class within the same genus as Ivory Gull. In \cref{table:genus_performance} we qualitatively gauge how misclassifications change when using \textit{nlg2choice} by measuring its \textit{genus-level} performance. We construct a species to genus mapping on the CUB200 dataset manually, then map the species prediction in previous results to genus. Simply put, we find that \textit{nlg2choice} makes more reasonable errors, switching to \textit{nlg2choice} incurs a genus-level performance improvement of \textbf{+17.91}, \textbf{+23.85}, \textbf{+23.59} for Qwen-2.5VL, Llama-3.2V-11B, and Intern3VL-8B.

\begin{figure}[t]
    \centering
    \setlength{\belowcaptionskip}{-15pt}
    \captionsetup{type=figure}
    \includegraphics[width=\linewidth]{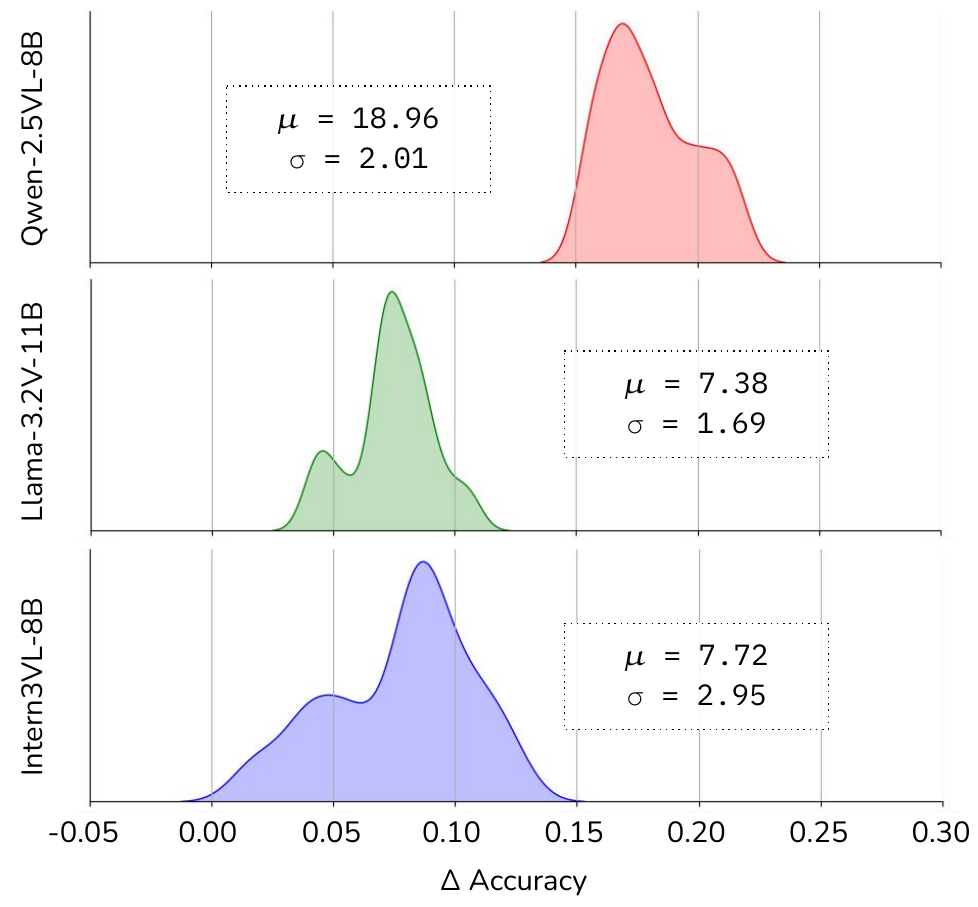}
    \vspace{-10pt}
    \captionof{figure}{\textbf{Difference in accuracies on the question-level between \textit{nlg2choice} and constrained decoding on CUB200}. Each distribution is made up of the performance difference when switching to \textit{nlg2choice} over 15 questions. \textit{nlg2choice} uses an open-ended prompt whereas constrained decoding uses an instruction to increase brevity (\texttt{"... Answer with species only."}).}
    \label{fig:statistical_significance}
\end{figure}%

\vspace{.3cm}
\noindent
\textbf{Improvements are statistically significant.} Rather than ask about the average performance of different methods under user variation, we wish to quantify how much performance is gained \textit{for a specific question by switching to nlg2choice}. We calculate this by taking the difference between the performance of constrained decoding and nlg2choice$_{open}$ for each question, then compute summary statistics over the set of questions. This forms a distribution of changes in accuracy, with significant results having a positive mean and 95\% confidence interval. The performance difference for the various settings in outlined in \cref{fig:statistical_significance}.

We find that for the proposed settings, switching from constrained decoding with an instruction to increase brevity (\texttt{"... Answer with species only."}) to nlg2choice$_{open}$ comes with an average $\mu = \mathbf{18.96}, \mathbf{7.38}, \mathbf{7.72}$ percentage increase in accuracy for Qwen-2.5VL, Llama-3.2V-11B, and Intern3VL-8B, respectively. Additionally, we find that these increases are within the bounds of statistical significance ($\sigma = \mathbf{2.01}, \mathbf{1.69}, \mathbf{2.95}$).

\vspace{.3cm}
\noindent
\textbf{Constrictive instructions reduce performance.} A common means of steering models to produce valid choices is through appending explicit instructions onto the instruction which enumerate the available choices, typically by means of lettering or enumeration. We wish to quantify the effect of adding this explicit instruction to the model. In \cref{table:constrictive_performance} we show the performance of constrained decoding and \textit{nlg2choice} under three variations of instruction steering: (1) \textit{no explicit choice steering} (\texttt{"What is the \{type\} of \{domain\} in this image?"}), (2) \textit{encouraging succinct prediction} (\texttt{"... Answer with \{type\} only"}), and (3) \textit{explicitly enumerating choices} (\texttt{"... Answer from \{choice\_list\}"}). 

\begin{table}[t]
\setlength{\belowcaptionskip}{-25pt}
\setlength{\tabcolsep}{7pt}
\begin{center}
{\footnotesize
\begin{tabular}{lcc}
\hline
 \\[-2ex]
\textbf{Method} & \textbf{Birds} & \textbf{Flowers} \\
 \\[-2ex]
\hline
\rowcolor{Gray} \hspace{3cm} && \\[-1.8ex]
\rowcolor{Gray}
\multicolumn{3}{l}{\texttt{"What is the \{type\} of \{domain\} in this image?"}} \\[.3ex]
 \\[-2ex]
Qwen-2.5VL & 0.92 & 2.53 \\ 
+ nlg2choice & \textbf{59.88 \cpar{58.96}} & \textbf{78.03 \cpar{75.5}} \\[.5ex] 
Llama-3.2-Vision & 7.67 & 27.50 \\ 
+ nlg2choice & 49.59 \cpar{41.92} & 62.24 \cpar{34.74} \\[.5ex] 
Intern3-VL & 2.28 & 5.01 \\ 
+ nlg2choice & 18.47 \cpar{16.19} & 37.5 \cpar{32.49} \\[.5ex]
 \\[-2ex]
\hline
\rowcolor{Gray} && \\[-1.8ex]
\rowcolor{Gray}
\multicolumn{3}{l}{\texttt{"... Answer with \{type\} only."}} \\[.3ex]
 \\[-2ex]
Qwen-2.5VL & 40.08 & 51.29 \\ 
+ nlg2choice & \textbf{47.72 \cpar{7.64}} & \textbf{63.22 \cpar{11.93}} \\[.5ex] 
Llama-3.2-Vision & 42.27 & 45.68 \\ 
+ nlg2choice & 46.96 \cpar{4.69} & 51.16 \cpar{5.48} \\[.5ex] 
Intern3-VL & 11.27 & 24.68 \\ 
+ nlg2choice & 13.81 \cpar{2.54} & 32.52 \cpar{7.84} \\[.5ex] 
 \\[-2ex]
\hline
\rowcolor{Gray} && \\[-1.8ex]
\rowcolor{Gray}
\multicolumn{3}{l}{\texttt{"... Answer from \{choice\_list\}."}} \\[.3ex]
 \\[-2ex]
Qwen-2.5VL & 38.71 & 69.19 \\ 
+ nlg2choice & 41.05 \cpar{2.34} & \textbf{72.48 \cpar{3.29}} \\[.5ex] 
Llama-3.2-Vision & 13.50 & 44.83 \\ 
+ nlg2choice & \textbf{46.68 \cpar{33.18}} & 66.03 \cpar{21.2} \\[.5ex] 
Intern3-VL & 22.34 & 35.24 \\ 
+ nlg2choice & 23.48 \cpar{1.14} & 35.73 \cpar{0.49} \\ 
 \\[-2ex]
\hline
\end{tabular}
\captionof{table}{\textbf{Classification performance across instruction templates with varying levels of constrictiveness.} The model name refers to default constrained decoding. The difference in performance is shown in parentheses. The best performing model and method for each input instruction is \textbf{bolded}. ``nlg2choice$_{open}$" corresponds to the top section of the table.}
\label{table:constrictive_performance}
}
\end{center}
\end{table}

We see that for every template, switching from constrained decoding to nlg2choice improves performance. The most stark setting is the most open-ended prompt, where we see an average increase of \textbf{+67.23}, \textbf{+38.33}, and \textbf{+24.34} for Qwen-2.5VL, Llama-3.2V, and Intern3-VL, respectively. Surprisingly, explicit choice instruction shows the weakest improvement for Qwen-2.5VL and Intern3-VL (\textbf{+2.82} and \textbf{+0.81}) but strong improvement for Llama-3.2V (\textbf{+27.19}). Otherwise, keeping in line with previous results, we find that Qwen-2.5VL is the greatest benefactor of performance.

\begin{table}[t]
\setlength{\belowcaptionskip}{-20pt}
\setlength{\tabcolsep}{7pt}
\begin{center}
{\footnotesize
\begin{tabular}{lcc}
\hline
 \\[-2ex]
\textbf{Method} & \textbf{Birds} & \textbf{Flowers} \\
 \\[-2ex]
\hline
\rowcolor{Gray} \hspace{3cm} && \\[-1.8ex]
\rowcolor{Gray}
\multicolumn{3}{l}{\texttt{"What is the \{type\} of \{domain\} in this image?"}} \\[.3ex]
 \\[-2ex]
Qwen-2.5VL & \textbf{59.88} & \textbf{78.03} \\ 
Llama-3.2-Vision & 49.59 & 62.24 \\ 
Intern3-VL & 18.47 & 37.50 \\
 \\[-2ex]
\hline
\rowcolor{softblue} && \\[-1.8ex]
\rowcolor{softblue}
\multicolumn{3}{l}{\texttt{... "Let's think step by step."} \cite{kojima2022large}} \\[.3ex]
 \\[-2ex]
Qwen-2.5VL & \textbf{54.93 \cpar{-4.95}} & \textbf{70.16 \cpar{-7.87}} \\ 
Llama-3.2-Vision & 39.87 \cpar{-9.72} & 39.22 \cpar{-23.02} \\ 
Intern3-VL & 19.11 \cpar{0.64} & 30.82 \cpar{-6.68} \\ 
 \\[-2ex]
\hline
\rowcolor{softblue} && \\[-1.8ex]
\rowcolor{softblue}
\multicolumn{3}{l}{\texttt{... "First,"} \cite{ahn2022can}} \\[.3ex]
 \\[-2ex]
Qwen-2.5VL & \textbf{53.51 \cpar{-6.37}} & \textbf{72.80 \cpar{-5.23}} \\ 
Llama-3.2-Vision & 41.84 \cpar{-7.75} & 45.31 \cpar{-16.93} \\ 
Intern3-VL & 19.65 \cpar{1.18} & 33.46 \cpar{-4.04} \\
 \\[-2ex]
\hline
\rowcolor{softblue} && \\[-1.8ex]
\rowcolor{softblue}
\multicolumn{3}{l}{\texttt{... "Let's solve this problem by splitting}} \\
\rowcolor{softblue}
\multicolumn{3}{l}{\texttt{it into steps."} \cite{reynolds2021prompt}} \\[.3ex]
 \\[-2ex]
Qwen-2.5VL & \textbf{54.69 \cpar{-5.19}} & \textbf{59.56 \cpar{-18.47}} \\ 
Llama-3.2-Vision & 40.11 \cpar{-9.48} & 38.08 \cpar{-24.16} \\ 
Intern3-VL & 19.68 \cpar{1.21} & 31.06 \cpar{-6.44} \\
 \\[-2ex]
\hline
\end{tabular}
\captionof{table}{\textbf{Classification performance when using assistant output prepending CoT methods.} All results are reported using \textit{nlg2choice} as an answer extraction method. The proposed setting is the first section in gray. Forced assistant prepending text is shown in \colorbox{softblue}{blue}.}
\label{table:cot_performance}
}
\end{center}
\end{table}

\vspace{.2cm}
\noindent
\textbf{CoT doesn't consistently improve performance.} Next, we ask how much performance can be improved by using CoT-style instructions. We investigate methods which induce reasoning by forcing the model to generate some initial tokens that are common to thought-out reasoning (eg. \texttt{"Let's think step by step"}), then letting the model freely respond. In \cref{table:cot_performance} we choose three of these methods and report their performance and difference between the default nlg2choice$_{open}$ setting.

In general, we see that these methods do not consistently improve performance, resulting in an average decrease of \textbf{-8.01}, \textbf{-15.18}, and \textbf{-2.36} across Qwen-2.5VL, Llama-3.2V, and Intern3-VL, respectively. The one exception to this rule is the CUB200 dataset with Intern3-VL, where we see an average increase of \textbf{+1.01}. 

\subsection{Zero-Shot Retrieval Performance}\label{sec:results:retrieval} 

\paragraph{Answer extraction is better than directly answering.} In \cref{table:retrieval_performance}, we report the performance of various models when viewing FGVC datasets as a one-vs-rest retrieval task. Like the classification setting, we find that when compared to exhaustive ``Yes/No" questioning, \textit{nlg2choice} generally improves with the exception of the Stanford Cars \cite{cars_dataset}, in which performance across all models decreases. We again find Qwen-2.5VL to receive the greatest improvement, with \textbf{+8.16} average mAP increase across the 5 datasets. Likewise, we find Intern3-VL receives the least improvement, with an average increase of \textbf{+4.43}.

Unlike classification, we do not find that changing the prompt to be open-ended (nlg2choice$_{open}$, ie. removing \texttt{"Answer from \{species\_list\}"} from the end of the input prompt) does not uniformly increase performance. However, we do still find that open-ended prompts generally perform better on average with \textit{nlg2choice} (\textbf{+8.16} vs. \textbf{+6.96}) and are less variable across all models.

% \paragraph{Truncation doesn't significantly change probabilities.}

% \paragraph{Numbers are difficult for probability generation.}

\begin{table}[t]
\setlength{\belowcaptionskip}{-20pt}
\setlength{\tabcolsep}{3.5pt}
\begin{center}
{\footnotesize
\begin{tabular}{lccccc}
\hline
 \\[-2ex]
\textbf{Method} & \textbf{Birds} & \textbf{Flowers} & \textbf{Aircrafts} & \textbf{Cars} & \textbf{Foods} \\
 \\[-2ex]
\hline
\rowcolor{Gray} &&&&& \\[-1.8ex]
\rowcolor{Gray}
\multicolumn{6}{l}{Baseline} \\[.3ex]
 \\[-2ex]
Random & 0.64 & 1.24 & 1.23 & 0.62 & 1.02 \\ 
 \\[-2ex]
\hline
\rowcolor{Gray} &&&&& \\[-1.8ex]
\rowcolor{Gray}
\multicolumn{6}{l}{Qwen-2.5VL-7B} \\[.3ex]
 \\[-2ex]
Yes/No & 51.01 & 61.19 & 51.07 & \textbf{77.58} & 82.67 \\[.5ex] 
nlg2choice & 57.79 & 82.56 & \textbf{65.14} & 68.61 & 84.30 \\ 
~ & \cpar{6.78} & \cpar{21.37} & \textbf{\cpar{14.07}} & \cpar{-8.97} & \cpar{1.63} \\[.5ex] 
nlg2choice$_{open}$ & \textbf{59.01} & \textbf{86.42} & 64.03 & 69.82 & \textbf{85.05} \\ 
~ & \textbf{\cpar{8.00}} & \textbf{\cpar{25.23}} & \cpar{12.96} & \cpar{-7.76} & \textbf{\cpar{2.38}} \\[.5ex] 
 \\[-2ex]
\hline
\rowcolor{Gray} &&&&& \\[-1.8ex]
\rowcolor{Gray}
\multicolumn{6}{l}{Llama-3.2-Vision-11B} \\[.3ex]
 \\[-2ex]
Yes/No & 41.24 & 63.59 & 34.24 & \textbf{66.03} & 78.81 \\[.5ex] 
nlg2choice & 51.14 & 68.64 & 39.39 & 56.46 & 86.73 \\ 
~ & \cpar{9.90} & \cpar{5.05} & \cpar{5.15} & \cpar{-9.57} & \cpar{7.92} \\[.5ex] 
nlg2choice$_{open}$ & \textbf{52.61} & \textbf{74.55} & \textbf{40.00} & 59.35 & \textbf{88.84} \\ 
~ & \textbf{\cpar{11.37}} & \textbf{\cpar{10.96}} & \textbf{\cpar{5.76}} & \cpar{-6.68} & \textbf{\cpar{10.03}} \\[.5ex]
 \\[-2ex]
\hline
\rowcolor{Gray} &&&&& \\[-1.8ex]
\rowcolor{Gray}
\multicolumn{6}{l}{Intern3VL-8B} \\[.3ex]
 \\[-2ex]
Yes/No & 15.22 & 38.32 & 17.93 & \textbf{31.83} & 61.46 \\[.5ex] 
nlg2choice & 26.97 & \textbf{45.13} & 20.47 & 25.81 & 66.55 \\ 
~ & \cpar{11.75} & \textbf{\cpar{6.81}} & \cpar{2.54} & \cpar{-6.02} & \cpar{5.09} \\[.5ex] 
nlg2choice$_{open}$ & \textbf{30.51} & 41.12 & \textbf{20.69} & 26.84 & \textbf{67.76} \\ 
~ & \textbf{\cpar{15.29}} & \cpar{2.80} & \textbf{\cpar{2.76}} & \cpar{-4.99} & \textbf{\cpar{6.30}} \\[.5ex]
 \\[-2ex]
\hline
\end{tabular}
\captionof{table}{\textbf{Performance across fine-grained visual classification (FGVC) benchmarks and architectures when interpreted as \textit{retrieval} tasks.} Numbers represent Mean Average Precision (mAP). Every query is a one-vs-rest task, eg. for CUB200 (Birds) the prior for a single PR curve is $1/200$.}
\label{table:retrieval_performance}
}
\end{center}
\end{table}

\subsection{Answer Extraction Performance} \label{sec:results:choice_ablation}

\begin{figure}[t]
    \centering
    \setlength{\belowcaptionskip}{-15pt}
    \captionsetup{type=figure}
    \includegraphics[width=\linewidth]{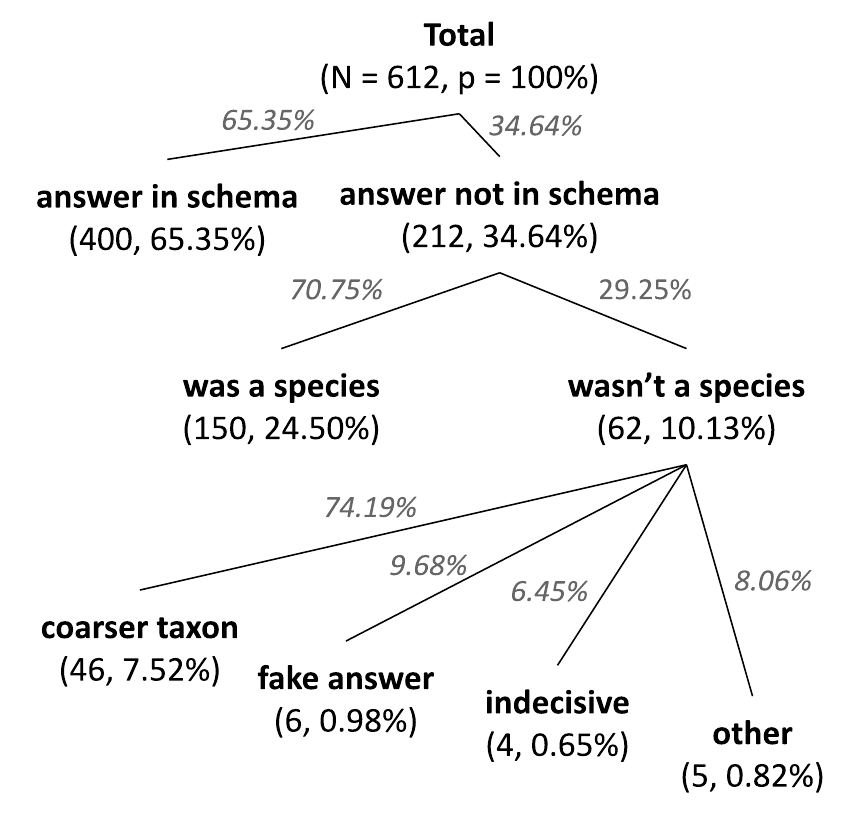}
    \vspace{-25pt}
    \captionof{figure}{\textbf{Breakdown of the small labeled set of natural language responses.} "Other" contains two main categories, "Refused to answer" and "No discernable information." Examples of these are given in \cref{sec:appendix:answer_extraction}.}
    \label{fig:extraction_dataset_breakdown}
\end{figure}

In \cref{sec:appendix:answer_extraction} we detail the process of labeling data for species extraction from free-form responses. First, labelers are shown responses from various models to a dataset instruction and image, then are instructed to highlight the first incidence of the predicted species within the generated text. Then, they are instructed to confirm whether the species highlighted matches the choice selection by the model, and in the failure case to indicate the species predicted or whether there was a schema failure. This process results in a set of common outcomes, ie. ``schema failure," ``no species predicted," ``answer=\texttt{\{species\}}". In \cref{fig:extraction_dataset_breakdown} we show the makeup of the labeled data in terms of these various outcomes. 
% For a full explanation of each node we refer to \cref{sec:appendix:answer_extraction_categories}.

\vspace{.3cm}
\noindent
\textbf{Models often respond out of schema.} Depicted in \cref{fig:extraction_dataset_breakdown}, we see that \textbf{34.64\%} of free-form responses by models have an answer within their text \textit{that does not occur within the dataset schema}. In other words, the worst-case scenario is that only \textbf{65.35\%} of examples are correctly classifiable. However, this turns out to not be the case due to the fact that a large percentage of these responses have real species within them (\textbf{70.75\%} of failures) which can be resolved to the correct answer, but that the predicted species itself cannot be easily put within the given schema. We hypothesize that many of these responses are grouped into the correct answer due to either genus-level answers in the dataset, eg. CUB200's labels include the genus \texttt{"Mockingbird"} which matches to \texttt{"Northern Mockingbird"} within the free-form text, or being a closely-related but technically different species, eg. \texttt{"Glaucous-winged Gull"} in schema versus \texttt{"Glaucous Gull"} in free-form text.

\vspace{.3cm}
\noindent
\textbf{Motivation of different methods.} In \cref{table:extraction_performance} we show the performance of main architectures in terms of extraction performance over various methods when sub-setting the data to examples where an answer within the schema is clearly apparent. We include two additional variations: (1) \textit{nlg2nlg}, the free-form response given the instruction to rephrase the MLLM answer and (2) \textit{nlg2nlg2choice}, the constrained decoding off one extra round of rephrasing. We include \textit{nlg2nlg} due to the fact that we wish to test the direct distance to the highlighted text created during labeling and include \textit{nlg2nlg2choice} to gauge the distance between the interpretation of that text into an answer. 

\vspace{.3cm}
\noindent
\textbf{LLMs are decent answer extractors out-of-the-box.} In the default nlg2choice$_{open}$ setting, we find high answer extraction success rates for Qwen-2.5VL (\textbf{97.93}) and Intern3-VL (\textbf{93.26}), but lower results for Llama-3.2V (\textbf{79.02}). We find that the raw text extracted by the models ("nlg2nlg$_{exact}$") seldom fits neatly into the dataset schema, having large performance drops (\textbf{-16.47}, \textbf{50.77}, \textbf{-25.91}). However, the output of nlg2nlg$_{exact}$ retains the relevant information from the original free-form text, as indicated by the nlg2nlg2choice scores which has much smaller performance differences (\textbf{+0.61}, \textbf{-2.72}, and \textbf{-4.39}). We hypothesize that the nlg2nlg text still contains artifacts such as \texttt{"The species is ..."} or scientific names, which lower exact match performance. 

% We provide an analysis of these cases, as well as a comparison of nlg2nlg to labeled spans in \cref{sec:appendix:text_span_extraction}.

\begin{table}[t]
\setlength{\belowcaptionskip}{-20pt}
\setlength{\tabcolsep}{4pt}
\begin{center}
{\scriptsize
\begin{tabular}{lcccc}
\hline
 \\[-2ex]
\textbf{Method} & \textbf{Qwen2.5VL} & \textbf{Llama3.2V} & \textbf{Intern3VL} & \textbf{Overall} \\
 \\[-2ex]
\hline
\rowcolor{Gray} &&&& \\[-1.8ex]
\rowcolor{Gray}
\multicolumn{5}{l}{xFinder \cite{yu2024xfinder}} \\[.3ex]
 \\[-2ex]
Qwen-1.5-0.5B \cite{qwen1.5} & \textbf{92.05} & \textbf{88.67} & 71.76 & \textbf{86.27} \\ 
Llama-3-8B-Instruct \cite{llama3} & 80.13 & 80.67 & \textbf{75.29} & 79.27 \\
 \\[-2ex]
\hline
\rowcolor{Gray} &&&& \\[-1.8ex]
\rowcolor{Gray}
\multicolumn{5}{l}{Qwen-2.5VL 8B} \\[.3ex]
 \\[-2ex]
nlg2choice & \textbf{98.01} & \textbf{98.67} & 96.47 & 97.93 \\[.5ex] 
nlg2nlg$_{exact}$ & 80.79 & 84.00 & 78.82 & 81.61 \\ 
~ & \cpar{-17.22} & \cpar{-14.67} & \cpar{-17.65} & \cpar{-16.32} \\[.5ex] 
nlg2nlg2choice & \textbf{98.01} & 97.33 & \textbf{100.00} & \textbf{98.19} \\ 
~ & \textbf{\cpar{0.00}} & \cpar{-1.34} & \textbf{\cpar{3.53}} & \textbf{\cpar{0.26}} \\
 \\[-2ex]
\hline
\rowcolor{Gray} &&&& \\[-1.8ex]
\rowcolor{Gray}
\multicolumn{5}{l}{Llama-3.2-Vision 11B} \\[.3ex]
 \\[-2ex]
nlg2choice & \textbf{77.48} & \textbf{82.00} & \textbf{76.47} & \textbf{79.02} \\[.5ex] 
nlg2nlg$_{exact}$ & 31.79 & 26.00 & 25.88 & 28.24 \\ 
~ & \cpar{-45.69} & \cpar{-56.00} & \cpar{-50.59} & \cpar{-50.78} \\[.5ex] 
nlg2nlg2choice & 76.82 & 80.00 & 70.59 & 76.68 \\ 
~ & \cpar{-0.66} & \cpar{-2.00} & \cpar{-5.88} & \cpar{-2.34} \\
 \\[-2ex]
\hline
\rowcolor{Gray} &&&& \\[-1.8ex]
\rowcolor{Gray}
\multicolumn{5}{l}{Intern3-VL 8B} \\[.3ex]
 \\[-2ex]
nlg2choice & \textbf{94.04} & \textbf{94.00} & \textbf{90.59} & \textbf{93.26} \\[.5ex] 
nlg2nlg$_{exact}$ & 59.60 & 78.67 & 62.35 & 67.62 \\ 
~ & \cpar{-34.44} & \cpar{-15.33} & \cpar{-28.24} & \cpar{-25.64} \\[.5ex] 
nlg2nlg2choice & 90.06 & 85.33 & \textbf{90.59} & 88.34 \\ 
~ & \cpar{-3.98} & \cpar{-8.67} & \textbf{\cpar{0.00}} & \cpar{-4.92} \\
 \\[-2ex]
\hline
\end{tabular}
% \captionof{table}{\textbf{Performance of various MLLMs across a small labeled set.} Accuracy is reported for each model over the full dataset (Overall), as well as the models making up the dataset. ``nlg2nlg" describes the raw ability of the model to extract the text and ``nlg2nlg2choice" is the information loss from the extracted text to the schema.}
\captionof{table}{\textbf{Performance of various MLLMs across a small labeled set.} Differences are measured from the ``nlg2choice`` row of each model.}
\label{table:extraction_performance}
}
\end{center}
\end{table}

\section{Limitations}

% (1) only medium size models were tested, (2) findings only hold within variance created by o3, (3) 

This work has three main limitations. First, our experiments were conducted primarily with medium-sized models (8B-11B) and we only used open-source LLMs from a few different families: Qwen, Llama, and Intern. Secondly, while nlg2choice avoids the need for specialized training data or architectural modifications and is more aligned with the true responses of the model, the method still requires significant computational resources to generate and process the textual descriptions. Lastly, while \textit{nlg2choice}
shows promise in classification, it still remains to be seen whether the approach scales to multi-label problems.

\section{Conclusion}

In this work, we presented nlg2choice, a simple approach for zero-shot fine-grained visual classification and retrieval that extracts fine-grained answers from MLLM free-form responses. Through experiments across multiple architectures and datasets, we demonstrated that nlg2choice consistently outperforms constrained decoding and exhaustive questioning methods in both classification accuracy and retrieval performance. We also validated the answer extraction directly through labeling predicted species in unstructured text, where we showed that models are capable of extraction their own answers without any labeled data. 

% Our findings highlight the importance of prompt engineering and the potential for improving fine-grained visual understanding without requiring specialized training data or architectural modifications. Overall, they suggest that answer extraction can serve as a practical tool for real-world applications where efficiency and simplicity are critical.

{
    \small
    \bibliographystyle{ieeenat_fullname}
    \bibliography{main}
}

\appendix

\section{Labeling Species Answer Extraction Data}\label{sec:appendix:answer_extraction}

% Data is generated by sampling 

% \paragraph{Labeling instructions.}  Our aim is to measure two capabilities: (1) to qualitatively gauge the quality of free-form generation with respect to species prediction (2)

\begin{figure*}[ht!]
    \centering
    \captionsetup{type=figure}
    \includegraphics[width=\linewidth]{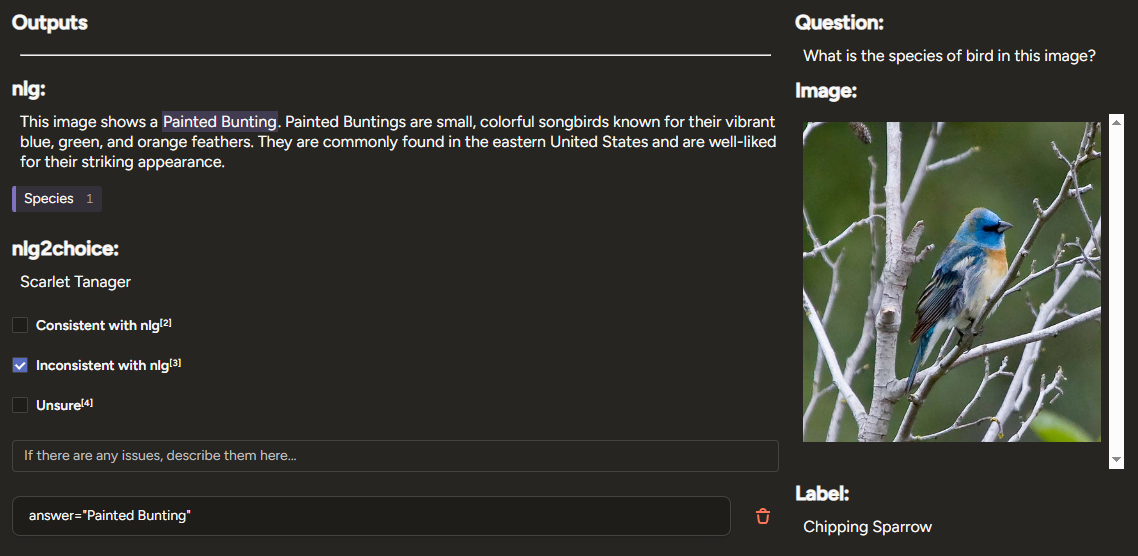}
    \vspace{-10pt}
    \captionof{figure}{\textbf{Labeling interface for efficacy of the answer extraction stage of \textit{nlg2choice}.} Labelers are asked to highlight the name predicted in free-form responses from various models and whether the name occurs easily within the dataset choice list. Above shows an interesting example: for the Chipping Sparrow on the right, the model freely responded with "Painted Bunting" and answer extraction made an additional mistake by predicting "Scarlet Tanager."}
    \label{fig:labeling_interface}
\end{figure*}

\paragraph{Motivation.}  For labeling answer extraction data, we prepare a user interface built with Label Studio which shown in \cref{fig:labeling_interface}. On the right contains information on the inputs to the model, specifically the question, image, and ground truth label. On the left is main interface, which consists a free-form response from a model to the example displayed on the right, as well as the choice predicted by the model via \textit{nlg2choice}. Labeling consists of two pieces: (1) identification and highlighting of text spans containing answers to the input question and (2) matching the highlighted text span to a class within the given schema.

\paragraph{Text span labeling.} For the text span labeling task, labelers read through the text to identify the first incidence of a predicted species. Before highlighting the species found within the text, the prediction is confirmed to be a real species through a brief internet search. If there are any issues with highlighting the predicted species, a text description of the issue is written in the input text box at the bottom and the example is skipped. 

\paragraph{Choice assignment.} Once the span is labeled, we next wish to either assign the span to a class within the dataset or tag the span as a schema failure. For expediency, the labelers are also equipped with the nlg2choice prediction and a tool for sorting most like class names with fuzzy matching score. When the nlg2choice predict matches the text span, the prediction is "Consistent with nlg" and the labelers check the button. Otherwise, "Inconsistent with nlg" is checked and a natural language explanation is put in the bottom text box. When the nlg2choice is inconsistent, ``answer=\texttt{\{species\}}" is put within that text box.

% First, labelers are shown responses from various models to a dataset instruction and image, then are instructed to highlight the first incidence of the predicted species within the generated text. Then, they are instructed to confirm whether the species highlighted matches the choice selection by the model, and in the failure case to indicate the species predicted or whether there was a schema failure. This process results in a set of common outcomes, ie. "schema failure," "no species predicted," "answer=\texttt{\{species\}}". In \cref{fig:extraction_dataset_breakdown} we show the makeup of the labeled data in terms of these various outcomes. For a full explanation of each node we refer to \cref{sec:appendix:answer_extraction_categories}.

% \section{Examples of Answer Extraction Categories}\label{sec:appendix:answer_extraction_categories}

\section{Zero-Shot 4-Way VQA Performance}\label{sec:results:4_way_vqa}

\begin{table}[h!]
\setlength{\belowcaptionskip}{-20pt}
\setlength{\tabcolsep}{3pt}
\begin{center}
{\footnotesize
\begin{tabular}{lccccc}
\hline
 \\[-2ex]
\textbf{Method} & \rotatebox{90}{\textbf{CUB200}} & \rotatebox{90}{\textbf{iNat-Animal}} & \rotatebox{90}{\textbf{iNat-Plant}} & \rotatebox{90}{\textbf{ImgNet-Animal}} & \rotatebox{90}{\textbf{ImgNet-Artifact}} \\
 \\[-2ex]
\hline
\rowcolor{Gray} &&&&& \\[-1.8ex]
\rowcolor{Gray}
\multicolumn{6}{l}{Qwen-2.5VL-7B} \\[.3ex]
 \\[-2ex]
 A/B/C/D & 65.50 & \textbf{41.33} & \textbf{41.61} & \textbf{85.20} & 80.01 \\[.5ex]
nlg2choice$_{open}$ & \textbf{67.43} & 40.19 & 40.03 & 84.03 & \textbf{81.81} \\ 
~ & \textbf{\cpar{1.93}} & \cpar{-1.14} & \cpar{-1.58} & \cpar{-1.17} & \textbf{\cpar{1.80}} \\
 \\[-2ex]
\hline
\rowcolor{Gray} &&&&& \\[-1.8ex]
\rowcolor{Gray}
\multicolumn{6}{l}{Llama-3.2-Vision-11B} \\[.3ex]
 \\[-2ex]
A/B/C/D & 65.52 & \textbf{32.44} & \textbf{31.88} & 79.93 & \textbf{75.89} \\[.5ex]
nlg2choice$_{open}$ & \textbf{68.79} & 31.52 & 31.57 & \textbf{81.39} & 75.29 \\ 
 & \textbf{\cpar{3.27}} & \cpar{-0.92} & \cpar{-0.31} & \textbf{\cpar{1.46}} & \cpar{-0.60} \\ 
 \\[-2ex]
\hline
\rowcolor{Gray} &&&&& \\[-1.8ex]
\rowcolor{Gray}
\multicolumn{6}{l}{Intern3VL-8B} \\[.3ex]
 \\[-2ex]
A/B/C/D & \textbf{50.52} & 35.40 & \textbf{36.39} & 77.50 & 69.41 \\[.5ex]
nlg2choice$_{open}$ & 47.30 & \textbf{35.86} & 35.64 & \textbf{81.91} & \textbf{70.02} \\ 
~ & \cpar{-3.22} & \textbf{\cpar{0.46}} & \cpar{-0.75} & \textbf{\cpar{4.41}} & \textbf{\cpar{0.61}} \\ 
 \\[-2ex]
\hline
\end{tabular}
\captionof{table}{\textbf{4-way VQA accuracy across fine-grained classifcation tasks.}}
\label{table:four_way_vqa}
}
\end{center}
\end{table}

\paragraph{Answer extraction does not consistently increase performance.} Following recent work \cite{tan2025vision}, we evalute \textit{nlg2choice} on popular fine-graiend classification datasets as a 4-way VQA task. iNat-Animal and iNat-Plant are the rows of iNat21 \cite{inat_dataset} when subset to rows which whose kingdom matches "Animalia" and "Plantae." ImgNet-Animal and ImgNet-Artifact are the rows of ImageNet-1K \cite{imagenet_dataset} subset to rows which have an anceter of "Animal" and "Artifact" in the WordNet hierarchy \cite{wordnet_dataset}. The choices for each example are from the top 3 SigLIP \cite{zhai2023sigmoid} predictions and the correct class. For the \textit{nlg2choice} prompt we use the same prompt as the previous work. We report the open-ended question performance in \cref{table:four_way_vqa} where we find that it slightly underperforms or has not effect on the lettering approach. Specifically, we find that accuracy changes by \textbf{-0.03}, \textbf{+0.58}, and \textbf{+0.30} on average for Qwen-2.5VL, Llama-3.2V, and Intern3VL, respectively.

\vspace{.3cm}
\noindent
\textbf{Decreasing the choice set improves performance.} We are also able to compare the 4-way VQA setting of CUB200 directly to the many-way prediction displayed in \cref{table:fgvc_performance}. We see that subsetting the full choice set to four choices incurs a performance improvement of \textbf{+7.55}, \textbf{+19.20}, and \textbf{+28.83} for Qwen-2.5VL, Llama-3.2V, and Intern3VL, respectively.

% \section{Chain-of-Thought Experiments}\label{sec:appendix:cot_experiments}

% \section{CUB200 Species to Genus}\label{sec:appendix:cub_species_to_genus}

% \section{Prompt Template Performance}\label{sec:appendix:prompt_performance}

% \section{Choice Prompt Ablation}\label{sec:appendix:nlg2choice_prompt_ablation}

\section{Prompt Variations}\label{sec:appendix:prompt_varations}

\begin{table}[h!]
\setlength{\belowcaptionskip}{-10pt}
\setlength{\tabcolsep}{3pt}
\begin{center}
{\scriptsize
\begin{tabular}{p{\linewidth}}
\hline
 \\[-2ex]
\texttt{"What \{ type \} is this \{ domain \}?"}

\texttt{"What is the \{ type \} of this \{ domain \}?"}

\texttt{"What is the \{ type \} of the \{ domain \}?"}

\texttt{"What is the \{ type \} of the \{ domain \} in this image?"}

\texttt{"What is the \{ type \} of the \{ domain \} in the image?"}

\texttt{"Identify this \{ domain \}'s \{ type \}."}

\texttt{"Name the \{ type \} shown in the image."}

\texttt{"Which \{ domain \} \{ type \} is pictured here?"}

\texttt{"Classify the \{ type \} of this \{ domain \}."}

\texttt{"What \{ domain \} \{ type \} does the photo depict?"}

\texttt{"Determine the \{ type \} of the \{ domain \} in view."}

\texttt{"Provide the common name of this \{ domain \}."}

\texttt{"To which \{ type \} does this \{ domain \} belong?"}

\texttt{"Label the \{ type \} of the \{ domain \} shown."}

\texttt{"Recognize and state this \{ domain \}'s \{ type \}."}

 % \\[-2ex]
 \\
\hline
\end{tabular}
\caption{\textbf{Base template variations generated by \cref{sec:method:prompt_variation}}}
\label{tab:prompt_hanna_recursive}
}
\end{center}
\end{table}

\begin{table}[h]
\setlength{\belowcaptionskip}{-20pt}
\setlength{\tabcolsep}{15pt}
\begin{center}
{\small
\begin{tabular}{lccc}
\hline
 \\[-2ex]
\textbf{Dataset} & \textbf{Type} & \textbf{Domain} \\
 \\[-2ex]
\hline
 \\[-2ex]
CUB200 & species & bird \\ 
Flowers & species & flower \\ 
Aircrafts & variant & aircraft \\ 
Cars & year, make, and model & car \\ 
Foods & name & food \\ 
NABirds & species & bird \\ 
iNat-Birds &  species & bird \\ 
 \\[-2ex]
\hline
\end{tabular}
\captionof{table}{\textbf{Dataset variables for filling out prompt templates.}}
}
\end{center}
\end{table}

% \begin{figure}[t]
%     \centering
%     \captionsetup{type=figure}
%     \includegraphics[width=\linewidth]{ICCV2025-Author-Kit-Feb/placeholders/truncated_prob_1.png}
%     \vspace{0pt}
%     \captionof{table}{\textbf{Number of forward passes needed for different methods in the retrieval setting of \textit{nlg2choice.}} "Full Prob" refers to the probability of the full class name $p(\texttt{[cname]})$ under the LLM, where \texttt{[cname]} is a common name of a class. "Yes"/"No" refers to the common $|\mathcal{X}|\cdot|\mathcal{Y}|$ method of calculating probability over all the classes $y_i \in \mathcal{Y}$. "Truncated Prob" refers to the process outlined in \cref{sec:method:retrieval}. The percentages in the "Yes"/"No" and "Truncated Prob" columns refer to the throughput increase over caculating the full sequence probabilities "Full Prob." Every setting for truncation is faster than "Yes"/"No" except for one benchmark - NABirds \cite{nabirds_dataset}.}
%     \label{fig:preview_combined}
% \end{figure}%

% \section{Text Span Extraction}\label{sec:appendix:text_span_extraction}

\end{document}